\documentclass[11pt,a4paper]{article}
\usepackage[hyperref]{emnlp2019}
\usepackage{times}
\usepackage{latexsym}
\usepackage{url}
\usepackage{amssymb}
\usepackage{amsmath}
\usepackage{algorithm}
\usepackage{algorithmic}
\usepackage{latexsym}
\usepackage{multirow}
\usepackage{mathtools}
\usepackage{esvect}
\usepackage{array}
\usepackage{arydshln}
\usepackage{tabularx}
\usepackage{graphicx}
\usepackage{pgfplots}
\pgfplotsset{compat=1.7}
\usepackage{booktabs}
\usepackage{graphicx}
\usepackage{comment}
\usepackage{CJKutf8}
\usepackage{color}
\usepackage{pbox}
\usepackage{array}

\newcommand{\red}[1]{\textcolor{red}{#1}}
\newcommand{\blue}[1]{\textcolor{blue}{#1}}
\newcolumntype{P}[1]{>{\centering\arraybackslash}p{#1}}
\newcommand\blfootnote[1]{%
	\begingroup
	\renewcommand\thefootnote{}\footnote{#1}%
	\addtocounter{footnote}{-1}%
	\endgroup
}

\aclfinalcopy 

\title{Aligning Cross-Lingual Entities with Multi-Aspect Information}

\author{Hsiu-Wei Yang$^{1}$\thanks{\hspace{0.1cm}~Equal contribution.}~, Yanyan Zou$^{2}$\footnotemark[1]~, Peng Shi$^{1}$\footnotemark[1]~, Wei Lu$^2$, Jimmy Lin$^1$, \and Xu Sun$^{3,4}$\\[0.5ex]
$^1$ David R. Cheriton School of Computer Science, University of Waterloo\\
$^2$ StatNLP Research Group, Singapore University of Technology and Design \\
$^3$ MOE Key Lab of Computational Linguistics, School of EECS, Peking University \\
$^4$ Center for Data Science, Beijing Institute of Big Data Research, Peking University \\[0.5ex]
{\tt \{h324yang,p8shi,jimmylin\}@uwaterloo.ca} \\ 
 \tt yanyan\_zou@mymail.sutd.edu.sg \\
  \tt luwei@sutd.edu.sg, \tt xusun@pku.edu.cn
}

\date{}

\begin{document}
\maketitle
\begin{abstract}
Multilingual knowledge graphs (KGs), such as YAGO and DBpedia, represent entities in different languages.
The task of cross-lingual entity alignment is to match entities in a source language with their counterparts in target languages.
In this work, we investigate embedding-based approaches to encode entities from multilingual KGs into the same vector space, where equivalent entities are close to each other.
Specifically, we apply graph convolutional networks (GCNs) to combine multi-aspect information of entities, including topological connections, relations, and attributes of entities, to learn entity embeddings.
To exploit the literal descriptions of entities expressed in different languages, we propose two uses of a pretrained multilingual BERT model to bridge cross-lingual gaps.
We further propose two strategies to integrate GCN-based and BERT-based modules to boost performance.
Extensive experiments on two benchmark datasets demonstrate that our method significantly outperforms existing systems.\blfootnote{This work has been accepted in 2019 Conference on Empirical Methods in Natural Language Processing and 9th International Joint Conference on Natural Language Processing as a full paper.}
\end{abstract}

\section{Introduction}

\begin{figure}[t]
\centering
\includegraphics[width=0.40\textwidth]{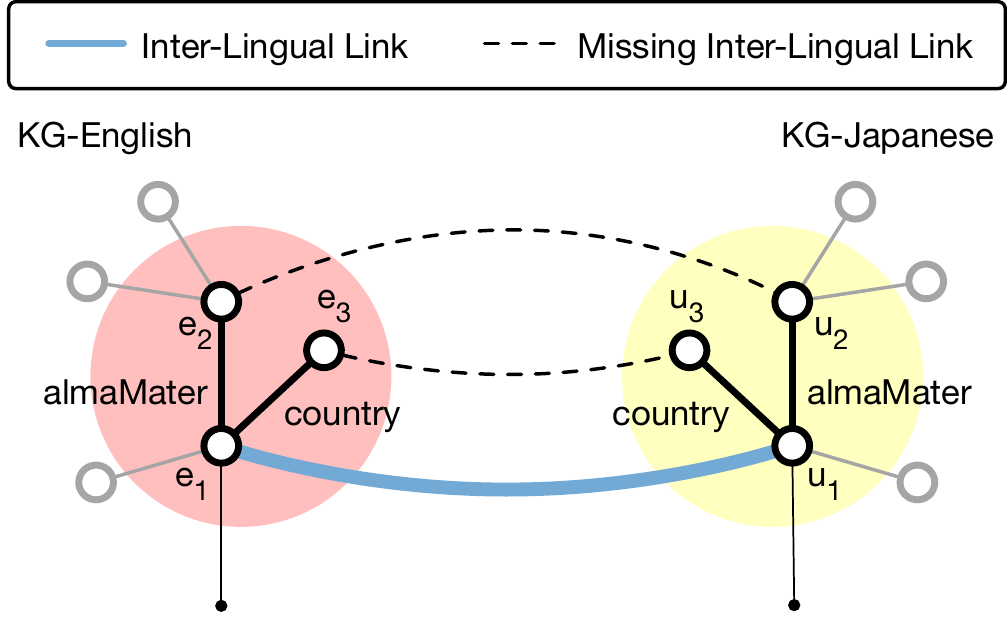}
\scalebox{0.6}{
\begin{tabular}{p{1.8cm}lp{2cm}p{3cm}}
\toprule
\multicolumn{2}{c}{English: University of Toronto}   & \multicolumn{2}{c}{Japanese: \begin{CJK}{UTF8}{min}トロント大学\end{CJK}} \\
\multicolumn{1}{c}{Attribute} & \multicolumn{1}{c}{Value} & \multicolumn{1}{c}{Attribute} & \multicolumn{1}{c}{Value} \\
\cmidrule(l){1-2} \cmidrule(l){3-4}
Name         & University of Toronto  &\begin{CJK}{UTF8}{min}大学名\end{CJK}        & \begin{CJK}{UTF8}{min}トロント大学\end{CJK}                    \\
Type         & Public University      & \begin{CJK}{UTF8}{min}学校種別\end{CJK}     &\begin{CJK}{UTF8}{min} 州立\end{CJK}                      \\
Found Date & 1827-03-15             & \begin{CJK}{UTF8}{min}創立年\end{CJK}      & 1827                    \\
Campus       & Ontario                & \begin{CJK}{UTF8}{min}キャンパス\end{CJK}    & \begin{CJK}{UTF8}{min}セントジョージ（トロント）\end{CJK}           \\
Former Name   & King's College         & \begin{CJK}{UTF8}{min}旧名\end{CJK}       & \begin{CJK}{UTF8}{min}キングスカレッジ\end{CJK}    \\     
\multicolumn{2}{c}{$\vdots$} & \multicolumn{2}{c}{$\vdots$}  \\
\bottomrule
\end{tabular}}
\scalebox{0.6}{
\begin{tabular}{p{5.5cm}p{5.5cm}}

\multicolumn{2}{c}{Descriptions}  \\
The University of Toronto is a public research university in Toronto, Ontario, Canada $\cdots$ & \begin{CJK}{UTF8}{min}トロント大学 は、オンタリオ州、トロントに本部を置くカナダの州立大学である $\cdots$\end{CJK} \\

\end{tabular}}
\caption{An example fragment of two KGs (in English and Japanese) connected by an inter-lingual link (ILL). In addition to the graph structures (top) consisting of entity nodes and typed relation edges, KGs also provide attributes and literal descriptions of entities (bottom).}
\label{fig:intro}
\end{figure}

A growing number of multilingual knowledge graphs (KGs) have been built, such as DBpedia \cite{bizer2009dbpedia}, YAGO \cite{suchanek2008yago,rebele2016yago}, and BabelNet \cite{navigli2012babelnet}, which typically represent real-world knowledge as separately-structured monolingual KGs.
Such KGs are connected via inter-lingual links (ILLs) that align entities with their counterparts in different languages, exemplified by Figure~\ref{fig:intro} (top).
Highly-integrated multilingual KGs contain useful knowledge that can benefit many knowledge-driven cross-lingual NLP tasks, such as machine translation \cite{moussallem2018machine} and cross-lingual named entity recognition \cite{darwish2013named}.
However, the coverage of ILLs among existing KGs is quite low \cite{chen2018co}:\ for example, less than 20\% of the entities in DBpedia are covered by ILLs.
The goal of cross-lingual entity alignment is to discover entities from different monolingual KGs that actually refer to the same real-world entities, i.e., discovering the missing ILLs.

Traditional methods for this task apply machine translation techniques to translate entity labels~\cite{spohr2011machine}.
The quality of alignments in the cross-lingual scenario heavily depends on the quality of the adopted translation systems.
In addition to entity labels, existing KGs also provide multi-aspect information of entities, including topological connections, relation types, attributes, and literal descriptions expressed in different languages~\cite{bizer2009dbpedia,xie2016representation}, as shown in Figure \ref{fig:intro} (bottom).
The key challenge of addressing such a task thus is how to better model and use provided multi-aspect information of entities to bridge cross-lingual gaps and find more equivalent entities (i.e., ILLs). 

Recently, embedding-based solutions~\cite{chen2017multi,sun2017cross,zhu2017iterative,wang2018cross,chen2018co} have been proposed to unify multilingual KGs into the same low-dimensional vector space where equivalent entities are close to each other.
Such methods only make use of one or two aspects of the aforementioned information.
For example, \citet{zhu2017iterative} relied only on topological features while \citet{sun2017cross} and \citet{wang2018cross} exploited both topological and attribute features.
\citet{chen2018co} proposed a co-training algorithm to combine topological features and literal descriptions of entities.
However, combining these multi-aspect information of entities (i.e., topological connections, relations and attributes, as well as literal descriptions)\ remains under-explored.

In this work, we propose a novel approach to learn cross-lingual entity embeddings by using all aforementioned aspects of information in KGs.
To be specific, we propose two variants of GCN-based models, namely \textsc{Man} and \textsc{Hman}, that incorporate multi-aspect features, including topological features, relation types, and attributes into cross-lingual entity embeddings.
To capture semantic relatedness of literal descriptions, we fine-tune the pretrained multilingual BERT model~\cite{devlin2019bert} to bridge cross-lingual gaps.
We design two strategies to combine GCN-based and BERT-based modules to make alignment decisions.
Experiments show that our method achieves new state-of-the-art results on two benchmark datasets. 
Source code for our models is publicly available at \url{https://github.com/h324yang/HMAN}.

\section{Problem Definition}
In a multilingual knowledge graph $\mathcal{G}$, we use $\mathcal{L}$ to denote the set of languages that $\mathcal{G}$ contains and 
$\mathcal{G}_i=\{E_i,R_i,A_i,V_i,D_i\}$ to represent the language-specific knowledge graph in language $L_i \in \mathcal{L}$.
$E_i$, $R_i$, $A_i$, $V_i$ and $D_i$ are sets of entities, relations, attributes, values of attributes, and literal descriptions, each of which portrays one aspect of an entity.
The graph $\mathcal{G}_i$ consists of relation triples $\left<h_i,r_i,t_i\right>$ and attribute triples $\left<h_i,a_i,v_i\right>$ such that $h_i,t_i \in E_i$, $r_i\in R_i$, $a_i \in A_i$ and $v_i \in V_i$.
Each entity is accompanied by a literal description consisting of a sequence of words in language $L_i$, e.g., $\left<h_i, d_{h,i}\right>$ and $\left<t_i, d_{t,i}\right>$, $d_{h,i},d_{t,i}\in D_i$.

Given two knowledge graphs $\mathcal{G}_{1}$ and $\mathcal{G}_{2}$ expressed in source language $L_1$ and target language $L_2$, respectively, there exists a set of pre-aligned ILLs $I\left(\mathcal{G}_{1}, \mathcal{G}_{2}\right) = \{\left(e, u\right) | e \in E_{1}, u\in E_{2}\}$ which can be considered training data. 
The task of cross-lingual entity alignment is to align entities in $\mathcal{G}_{1}$ with their cross-lingual counterparts in $\mathcal{G}_{2}$, i.e., discover missing ILLs.

\section{Proposed Approach}
In this section, we first introduce two GCN-based models, namely \textsc{Man} and \textsc{Hman}, that learn entity embeddings from the graph structures.
Second, we discuss two uses of a multilingual pretrained BERT model to learn cross-lingual embeddings of entity descriptions:\ \textsc{PointwiseBert} and \textsc{PairwiseBert}.
Finally, we investigate two strategies to integrate the GCN-based and the BERT-based modules.

\subsection{Cross-Lingual Graph Embeddings}
\label{sec:gcn}
Graph convolutional networks (GCNs) \cite{kipf2016semi} are variants of convolutional networks that have proven effective in capturing information from graph structures, 
such as dependency graphs \cite{dcgcnforgraph2seq19guo}, abstract meaning representation graphs \cite{aggcn19guo}, and knowledge graphs \cite{wang2018cross}.
In practice, multi-layer GCNs are stacked to collect evidence from multi-hop neighbors.
Formally, the $l$-th GCN layer takes as input feature representations $H^{(l-1)}$ and outputs $H^{(l)}$:
\begin{align}
H^{(l)} = \phi\left( \tilde{D}^{-\frac{1}{2}}\tilde{A}\tilde{D}^{-\frac{1}{2}}H^{(l-1)}W^{(l)} \right)
\label{eq:gcn-layer}
\end{align}
where $\tilde{A} = A + I$ is the adjacency matrix, $I$ is the identity matrix, $\tilde{D}$ is the diagonal node degree matrix of $\tilde{A}$, $\phi(\cdot)$ is ReLU function, and $W^{(l)}$ represents learnable parameters in the $l$-{th} layer. 
$H^{(0)}$ is the initial input.

GCNs can iteratively update the representation of each entity node via a propagation mechanism through the graph.
Inspired by previous studies \cite{zhang2018graph,wang2018cross}, we also adopt GCNs in this work to collect evidence from multilingual KG structures and to learn cross-lingual embeddings of entities.
The primary assumptions are:\ (1) equivalent entities tend to be neighbored by equivalent entities via the same types of relations; (2) equivalent entities tend to share similar or even the same attributes.

\smallskip \noindent \textbf{Multi-Aspect Entity Features.} 
Existing KGs \cite{bizer2009dbpedia,suchanek2008yago,rebele2016yago} provide multi-aspect information of entities.
In this section, we mainly focus on the following three aspects:\ topological connections, relations, and attributes.
The key challenge is how to utilize the provided features to learn better embeddings of entities.
We discuss how we construct raw features for the three aspects, which are then fed as inputs to our model.
We use $X_t$, $X_r$ and $X_a$ to denote the topological connection, relation, and attribute features, individually.

The topological features contain rich neighborhood proximity information of entities, which can be captured by multi-layer GCNs.
As in \citet{wang2018cross}, we set the initial topological features to $X_t = I$, i.e., an identity matrix serving as index vectors (also known as the featureless setting), so that the GCN can learn the representations of corresponding entities.

In addition, we also consider the relation and attribute features.
As shown in Figure \ref{fig:intro}, the connected relations and attributes of two equivalent entities, e.g., ``\emph{University of Toronto}'' (English) and ``\begin{CJK}{UTF8}{min}トロント大学\end{CJK}'' (Japanese), have a lot of overlap, which can benefit cross-lingual entity alignment.
Specifically, they share the same relation types, e.g., ``{country}'' and ``{almaMater}'', and some attributes, e.g., ``{foundDate}'' and ``\begin{CJK}{UTF8}{min}創立年\end{CJK}''.
To capture relation information, \citet{schlichtkrull2018modeling} proposed RGCN with relation-wise parameters.
However, with respect to this task, existing KGs typically contain thousands of relation types but few pre-aligned ILLs.
Directly adopting RGCN may introduce too many parameters for the limited training data and thus cause overfitting.
\citet{wang2018cross} instead simply used the unlabeled GCNs \cite{kipf2016semi} with two proposed measures (i.e., functionality and inverse functionality) to encode the information of relations into the adjacency matrix.
They also considered attributes as input features in their architecture.
However, this approach may lose information about relation types.
Therefore, we regard relations and attributes of entities as bag-of-words features to explicitly model these two aspects.
Specifically, we construct count-based \textit{N-hot} vectors $X_r$ and $X_a$ for these two aspects of features, respectively, where the $(i,j)$ entry is the count of the $j$-th relation~(attribute) for the corresponding entity $e_i$. 
Note that we only consider the top-$F$ most frequent relations and attributes to avoid data sparsity issues.
Thus, for each entity, both of its relation and attribute features are $F$-dimensional vectors. 
 
\smallskip \noindent \textbf{\textsc{Man}.} 
Inspired by \citet{wang2018cross}, we propose the \emph{Multi-Aspect Alignment Network} (\textsc{Man}) to capture the three aspects of entity features. 
Specifically, three $l$-layer GCNs take as inputs the triple-aspect features (i.e., $X_t$, $X_r$, and $X_a$) and produce the representations $H_t^{(l)}$, $H_r^{(l)}$, and $H_a^{(l)}$ according to Equation~\ref{eq:gcn-layer}, respectively. 
Finally, the multi-aspect entity embedding is:
\begin{align}
H_m = [{H_t^{(l)}}\oplus{H_a^{(l)}}\oplus{H_r^{(l)}}]
\end{align}
\noindent where $\oplus$ denotes vector concatenation.
$H_m$ can then feed into alignment decisions.

Such fusion through concatenation is also known as \emph{Scoring Level Fusion}, which has been proven simple but effective for capturing multi-modal semantics \cite{bruni2014multimodal, kiela2014learning, collell2017imagined}.
It is worth noting that the main differences between \textsc{Man} and the work of \citet{wang2018cross} are two fold:\
First, we use the same approach as in \citet{kipf2016semi} to construct the adjacency matrix, while \citet{wang2018cross} designed a new connectivity matrix as the adjacency matrix for the GCNs.
Second, \textsc{Man} explicitly regards the relation type features as model input, while \citet{wang2018cross} incorporated such relation information into the connectivity matrix.

\begin{figure}[t]
\centering
\includegraphics[width=0.40\textwidth]{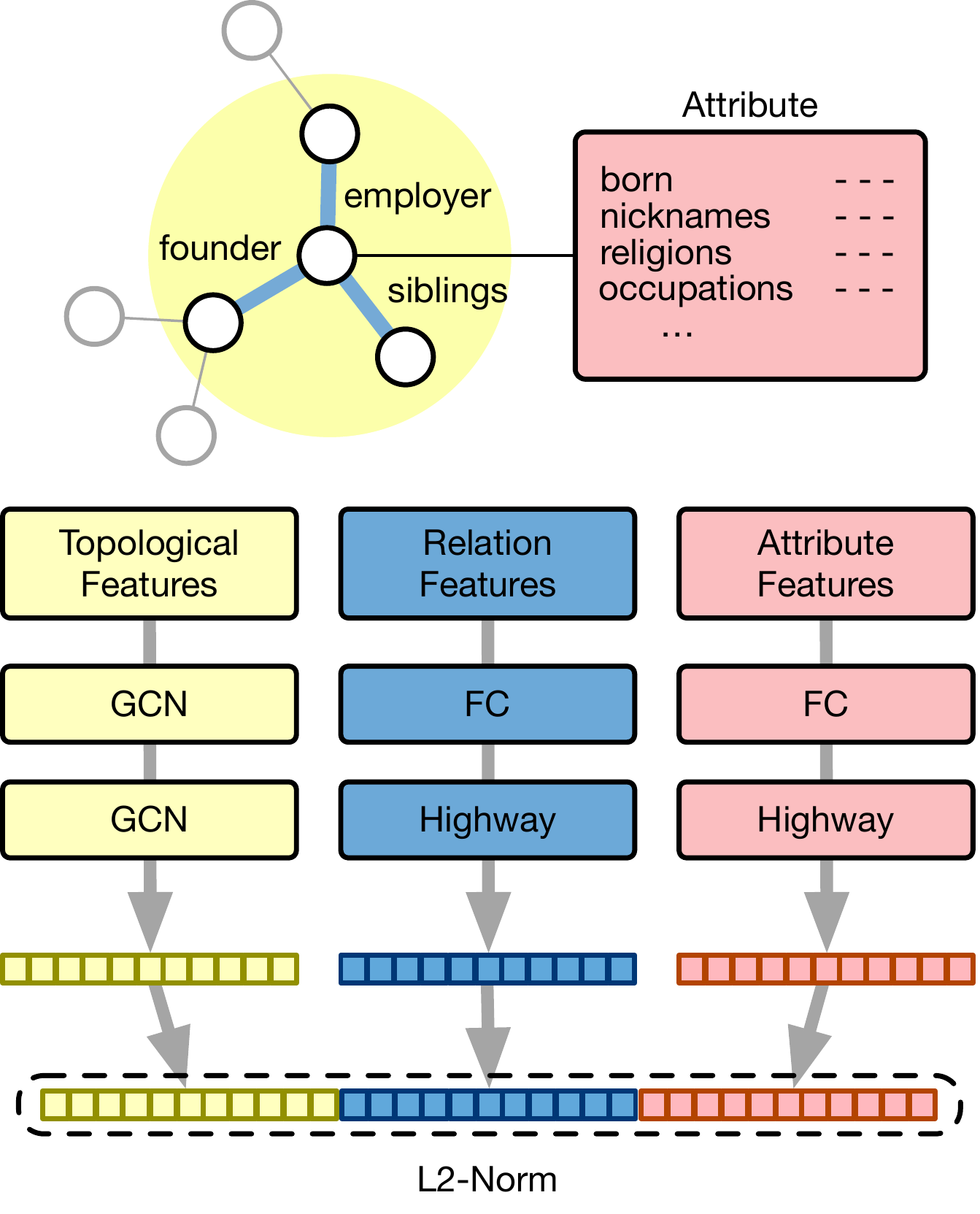}
\caption{Architecture of \textsc{Hman}.}
\label{fig:NMGCN}
\end{figure}

\begin{figure*}[t]
\centering
\includegraphics[width=0.85\textwidth]{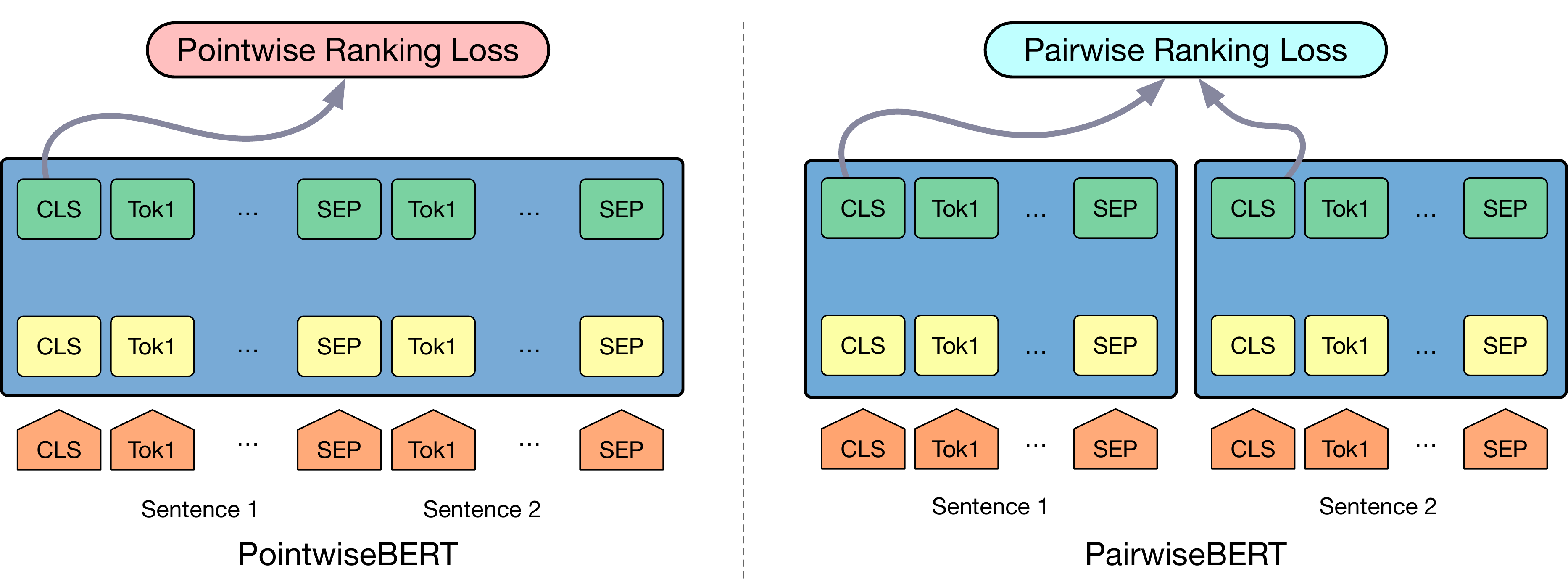}
\caption{Architecture overview of \textsc{PointwiseBert} (left) and \textsc{PairwiseBert} (right).}
\label{fig:pairwiseBERT}
\end{figure*}

\smallskip \noindent \textbf{\textsc{Hman}.}
Note that \textsc{Man} propagates relation and attribute information through the graph structure.
However, for aligning a pair of entities, we observe that considering the relations and attributes of neighboring entities, besides their own ones, may introduce noise.
Merely focusing on relation and attribute features of the current entity could be a better choice.
Thus, we propose the \emph{Hybrid Multi-Aspect Alignment Network} (\textsc{Hman}) to better model such diverse features, shown in Figure~\ref{fig:NMGCN}.
Similar to \textsc{Man}, we still leverage the \mbox{$l$-th} layer of a GCN to obtain topological embeddings $H_t^{(l)}$, but exploit feedforward neural networks to obtain the embeddings with respect to relations and attributes.
The feedforward neural networks consist of one fully-connected (FC) layer and a highway network layer~\cite{srivastava2015training}. 
The reason we use highway networks is consistent with the conclusions of~\citet{mudgal2018deep}, who conducted a design space exploration of neural models for entity matching and found that highway networks are generally better than FC layers in convergence speed and effectiveness.

Formally, these feedforward neural networks are defined as:
\begin{align}
S_{f}  &= \phi (W_f^{(1)}X_f+b_f^{(1)}) \notag \\ 
T_f  &= \sigma(W_f^t S_{f} + b_f^t) \\ 
G_f &= \phi(W_f^{(2)}S_{f} + b_f^{(2)})\cdot T_f + S_{f}\cdot(1-T_f) \notag
\end{align}
where $f \in \{r,a\}$ and $X_f$ refer to one specific aspect (i.e., relation or attribute) and the corresponding raw features, respectively, 
$W_f^{(1,2,t)}$ and $b_f^{(1,2,t)}$ are model parameters,
$\phi(\cdot)$ is ReLU function, and $\sigma(\cdot)$ is sigmoid function.
Accordingly, we obtain the hybrid multi-aspect entity embedding $H_y = [H_t^{(l)}\oplus G_r\oplus G_a]$, to which $\ell_2$ normalization is further applied. 

\smallskip
\noindent
\textbf{Model Objective.}
\label{learning_obj}
Given two knowledge graphs, $\mathcal{G}_{1}$ and $\mathcal{G}_{2}$, and a set of pre-aligned entity pairs $I\left(\mathcal{G}_{1}, \mathcal{G}_{2}\right)$ as training data, 
our model is trained in a supervised fashion.
During the training phase, the goal is to embed cross-lingual entities into the same low-dimensional vector space where equivalent entities are close to each other.
Following \citet{wang2018cross}, our margin-based ranking loss function is defined as:
\begin{align}
J =  &\sum_{(e_{1}^{}, e_{2}^{})\in I}\sum_{(e^{'}_{1}, e^{'}_{2})\in I^{'}} [ \rho(h_{e_{1}^{}}, h_{e_{2}^{}}) + \beta \notag \\
&- \rho(h_{e^{'}_{1}}, h_{e^{'}_{2}})]_{+}
\label{eq:loss}
\end{align}
where $[x]_{+} = \max\{0, x\}$, $I^{'}$ denotes the set of negative entity alignment pairs constructed by corrupting the gold pair $(e_{1}^{}, e_{2}^{})\in I$.
Specifically, we replace $e_{1}^{}$ or $e_{2}^{}$ with a randomly-chosen entity in ${E}_{1}$ or ${E}_{2}$.
$\rho(x,y)$ is the $\ell_1$ distance function, and $\beta > 0$ is the margin hyperparameter separating positive and negative pairs.

\subsection{Cross-Lingual Textual Embeddings}
\label{sec:pairwiseBERT}
Existing multilingual KGs \cite{bizer2009dbpedia,navigli2012babelnet,rebele2016yago} also provide literal descriptions of entities expressed in different languages and contain detailed semantic information about the entities.
The key observation is that literal descriptions of equivalent entities are semantically close to each other.
However, it is non-trivial to directly measure the semantic relatedness of two entities' descriptions, since they are expressed in different languages.

Recently, Bidirectional Encoder Representations from Transformer (BERT)~\cite{devlin2019bert} has advanced the state-of-the-art in various NLP tasks by heavily exploiting pretraining based on language modeling.
Of special interest is the multilingual variant, which was trained with Wikipedia dumps of 104 languages.
The spirit of BERT in the multilingual scenario is to project words or sentences from different languages into the same semantic space.
This aligns well with our objective---bridging gaps between descriptions written in different languages.
Therefore, we propose two methods for applying multilingual BERT, \textsc{PointwiseBert} and \textsc{PairwiseBert}, to help make alignment decisions.

\smallskip \noindent
\textbf{\textsc{PointwiseBert}.}
A simple choice is to follow the basic design of BERT and formulate the entity alignment task as a text matching task.
For two entities $e_1$ and $e_2$ from two KGs in $L_1$ and $L_2$, denoting source language and target language, respectively, their textual descriptions are $d_1$ and $d_2$, consisting of word sequences in two languages.
The model takes as inputs [CLS] \textit{$d_1$} [SEP] \textit{$d_2$} [SEP], where [CLS] is the special classification token, from which the final hidden state is used as the sequence representation, and [SEP] is the special token for separating token sequences, and produces the probability of classifying the pair as equivalent entities.
The probability is then used to rank all candidate entity pairs, i.e., ranking score.
We denote this model as \emph{\textsc{PointwiseBert}}, shown in Figure~\ref{fig:pairwiseBERT} (left).

This approach is computationally expensive, since for each entity we need to consider all candidate entities in the target language.
One solution, inspired by the work of \citet{shi2019simple}, is to reduce the search space for each entity with a \emph{reranking strategy} (see Section \ref{sec:strategy}).

\smallskip \noindent
\textbf{\textsc{PairwiseBert}.}
Due to the heavy computational cost of \textsc{PointwiseBert}, semantic matching between all entity pairs is very expensive.
Instead of producing ranking scores for description pairs, we propose \textsc{PairwiseBert} to encode the entity literal descriptions as cross-lingual textual embeddings, where distances between entity pairs can be directly measured using these embeddings.

The \textsc{PairwiseBert} model consists of two components, each of which takes as input the description of one entity (from the source or target language), as depicted in Figure~\ref{fig:pairwiseBERT} (right).
Specifically, the input is designed as [CLS] \textit{$d_1(d_2)$} [SEP], which is then fed into \textsc{PairwiseBert} for contextual encoding.
We select the hidden state of [CLS] as the textual embedding of the entity description for training and inference.
To bring the textual embeddings of cross-lingual entity descriptions into the same vector space,
a similar ranking loss function as in Equation \ref{eq:loss} is used.

\subsection{Integration Strategy}
\label{sec:strategy}
Sections~\ref{sec:gcn} and \ref{sec:pairwiseBERT} introduce two modules that separately collect evidence from knowledge graph structures and the literal descriptions of entities, namely graph and textual embeddings.
In this section, we investigate two strategies to integrate these two modules to further boost performance.

\smallskip \noindent
\textbf{Reranking.} 
As mentioned in Section \ref{sec:pairwiseBERT}, the \textsc{PointwiseBert} model takes as input the concatenation of two descriptions for each candidate--entity pair, where conceptually we must process every possible pair in the training set.
Such a setting would be cost prohibitive computationally.

One way to reduce the cost of \textsc{PointwiseBert} would be to ignore candidate pairs that are unlikely to be aligned. 
\citet{rao2016noise} showed that uncertainty-based sampling can provide extra improvements in ranking.
Following this idea, the GCN-based models (i.e., \textsc{Man} and \textsc{Hman}) are used to generate a candidate pool whose size is much smaller than the entire universe of entities.
Specifically, GCN-based models provide top-$q$ candidates of target entities for each source entity (where $q$ is a hyperparameter).
Then, the \textsc{PointwiseBert} model produces a ranking score for each candidate--entity pair in the pool to further rerank the candidates.
However, the weakness of such a reranking strategy is that performance is bounded by the quality of (potentially limited) candidates produced by \textsc{Man} or \textsc{Hman}.

\smallskip \noindent
\textbf{Weighted Concatenation.}
With the textual embeddings learned by \textsc{PairwiseBert} denoted as $H^B$ and graph embeddings denoted as $H^G$,
a simple way to combine the two modules is by weighted concatenation:
\begin{align}
H^C = \tau \cdot H^{G} \oplus (1-\tau)\cdot H^{B}
\label{equ:concate}
\end{align}
where $H^G$ is the graph embeddings learned by either \textsc{Man} or \textsc{Hman}, 
and $\tau$ is a factor to balance the contribution of each source (where $\tau$ is a hyperparameter).

\subsection{Entity Alignment}
After we obtain the embeddings of entities, we leverage $\ell_1$ distance to measure the distance between candidate--entity pairs. 
A small distance reflects a high probability for an entity pair to be aligned as equivalent entities.
To be specific, with respect to the reranking strategy, we select the target entities that have the smallest distances to a source entity in the vector space learned by \textsc{Man} or \textsc{Hman} as its candidates. 
For weighted concatenation, we employ the $\ell_1$ distance of the representations of a pair derived from the concatenated embedding, i.e., $H^C$, as the ranking score. 

\section{Experiments}
\label{sec:experiment}
\subsection{Datasets and Settings}
We evaluate our methods over two benchmark datasets:\ DBP15K and DBP100K \cite{sun2017cross}.
Table \ref{tab:data_stats} outlines the statistics of both datasets, which contain 15,000 and 100,000 ILLs, respectively.
Both are divided into three subsets:\ Chinese-English (ZH-EN), Japanese-English (JA-EN), and French-English (FR-EN).

Following previous work \cite{sun2017cross,wang2018cross}, we adopt the same split settings in our experiments, where 30\% of the ILLs are used as training and the remaining 70\% for evaluation.
\emph{Hits@k} is used as the evaluation metric \cite{bordes2013translating,sun2017cross,wang2018cross}, which measures the proportion of correctly aligned entities ranked in the top-$k$ candidates, 
and results in both directions, e.g., ZH-EN and EN-ZH, are reported.

In all our experiments, we employ two-layer GCNs and the top 1000 (i.e., $F$=1000) most frequent relation types and attributes are included to build the $N$-hot feature vectors.
For the \textsc{Man} model, we set the dimensionality of topological, relation, and attribute embeddings to 200, 100, and 100, respectively.
When training \textsc{Hman}, the hyperparameters are dependent on the dataset sizes due to GPU memory limitations.
For DBP15K, we set the dimensionality of topological embeddings, relation embeddings, and attribute embeddings to 200, 100, and 100, respectively.
For DBP100K, the dimensionalities are set to 100, 50, and 50, respectively.
We adopt SGD to update parameters and the numbers of epochs are set to 2,000 and 50,000 for \textsc{Man} and \textsc{Hman}, respectively.
The margin $\beta$ in the loss function is set to 3.
The balance factor $\tau$ is determined by grid search, which shows that the best performance lies in the range from 0.8 to 0.7.
For simplicity, $\tau$ is set to 0.8 in all associated experiments.
Multilingual BERT-base models with 768 hidden units are used in \textsc{PointwiseBert} and \textsc{PairwiseBert}.
We additionally append one more FC layer to the representation of [CLS] and reduce the dimensionality to 300.
Both BERT models are fine-tuned using the Adam optimizer. 

\begin{table}[t]
	\centering
	\scalebox{0.6}{
		\begin{tabular}{llcccccccccc}
			\toprule
			\multicolumn{2}{c}{\multirow{2}{*}{Datasets}}   &    \multicolumn{5}{c}{DBP15K}    \\
			\cmidrule(l){3-7} 
			&    & Entities & Rel. & Attr. & Rel.triples & Attr.triples  \\ 
			\midrule	
			\multirow{2}{*}{ZH-EN}   & Chinese  & 66,469  & 2,830 & 8,113 & 153,929 & 379,684        \\ 
			& English  & 98,125  &2,317 & 7,173   & 237,674  & 567,755      \\ 
			\multirow{2}{*}{JA-EN}  & Japanese & 65,744   & 2,043      & 5,882      &164,373      & 354,619  \\ 
			& English  &95,680   & 2,096      & 6,066      & 233,319      & 497,230    \\ 
			\multirow{2}{*}{FR-EN}  & French   & 66,858   &1,379      & 4,547       &192,191     & 528,665  \\ 
			& English  &105,889  & 2,209      & 6,422      &278,590     &576,543    \\
			\midrule
			\multicolumn{2}{c}{\multirow{2}{*}{Datasets}}    & \multicolumn{5}{c}{DBP100K} \\\cmidrule(l){3-7} 
			&     & Entities & Rel. & Attr. & Rel.triples & Attr.triples  \\ 
			\midrule	
			\multirow{2}{*}{ZH-EN}   & Chinese  & 106,517 & 4,431  & 16,152  & 329,890  & 1,404,615      \\ 
			& English  & 185,022   & 3,519      & 14,459      & 453,248      & 1,902,725      \\ 
			\multirow{2}{*}{JA-EN}  & Japanese  & 117,836   & 2,888      & 12,305      & 413,558      & 1,474,721 \\ 
			& English & 118,570   & 2,631      & 13,238      & 494,087      & 1,738,803    \\ 
			\multirow{2}{*}{FR-EN}  & French   & 105,724   & 1,775      & 8,029       & 409,399      & 1,361,509  \\ 
			& English  & 107,231   & 2,504      & 13,170      & 513,382      & 1,957,813    \\
			\bottomrule
	\end{tabular}}
	\caption{Statistics of DBP15K and DBP100K. Rel.\ and Attr.\ stand for relations and attributes, respectively.}
	\label{tab:data_stats}
\end{table}

\begin{table*}[ht]
\centering
\scalebox{0.79}{
	\begin{tabular}{lccccccc}
		\toprule
		\multicolumn{1}{c}{\multirow{3}{*}{Model}}          &    \multicolumn{1}{c}{ ZH $\to$ EN  }   &  EN$\to$ ZH &   JA $\to$ EN       & EN$\to$ JA &  FR $\to$ EN & EN$\to$ FR    \\
		\cmidrule(l){2-2} \cmidrule(l){3-3}  \cmidrule(l){4-4}  \cmidrule(l){5-5} \cmidrule(l){6-6} \cmidrule(l){7-7} 	 
		& @1 {\color{white}{.}}@10 @50 & @1 {\color{white}{.}}@10 @50 & @1 {\color{white}{.}}@10 @50 & @1 {\color{white}{.}}@10 @50  & @1 {\color{white}{.}}@10 @50 & @1 {\color{white}{.}}@10 @50 \\
		\midrule	
		&\multicolumn{7}{c}{{DBP15K}} \\
		\cmidrule(l){2-7}
		\citet{hao2016joint}      & 21.2 42.7 56.7    & 19.5 39.3 53.2  & 18.9 39.9 54.2   &  17.8  38.4 52.4   &  15.3 38.8 56.5 &  14.6 37.2 54.0       \\
		\citet{chen2017multilingual}  & 30.8 61.4 79.1 & 24.7 52.4 70.4   & 27.8 57.4 75.9   &  23.7 49.9 67.9 & 24.4 55.5 74.4      & 21.2 50.6  69.9    \\
		\citet{sun2017cross}       & 41.1 74.4 88.9  &  40.1 71.0 86.1  & 36.2 68.5 85.3  &  38.3 67.2 82.6 & 32.3 66.6 83.1   & 32.9 65.9 82.3     \\
		\citet{wang2018cross}     & 41.2 74.3 86.2 & 36.4 69.9 82.4 & 39.9 74.4 86.1 & 38.4 71.8 83.7   & 37.2 74.4 86.7   & 36.7 73.0 86.3    \\
		\hdashline
		\textsc{Man} & 46.0 79.4 90.0 & 41.5 75.6 88.3  & 44.6 78.8 90.0 & 43.0  77.1 88.7  & 43.1 79.7 91.7 & 42.1 79.1 90.9  \\
		\textsc{Man} w/o \textsc{te} & 21.5 55.0 79.4 & 20.2 53.6 78.8 & 15.0 44.0 69.9 &14.3 44.0 70.6 & 10.2  34.5 59.5 &10.8  35.2 60.3 \\
		\textsc{Man} w/o \textsc{re}   & 45.6 79.1 89.5 &  41.1 75.0 87.3  & 44.2 78.7 89.8 & 43.0 76.9 88.1  & 42.8  79.7 91.4  & 42.1 78.9 90.6 \\
		\textsc{Man} w/o \textsc{ae}    & 43.7 77.1 87.8 &39.2  72.9 85.5  & 43.2 77.6 88.4 & 41.2  74.9  86.6  & 42.9 79.6 91.0 &  41.5 78.9 90.5 \\
		
		\hdashline  
		\textsc{Hman}    & \textbf{56.2} \textbf{85.1} \textbf{93.4}    &\textbf{53.7}  \textbf{83.4} \textbf{92.5}    & \textbf{56.7} \textbf{86.9} \textbf{94.5}    &  \textbf{56.5} \textbf{86.6} \textbf{94.6} & \textbf{54.0} \textbf{87.1} \textbf{95.0}   & \textbf{54.3}  \textbf{86.7}  \textbf{95.1} \\
		\textsc{Hman} w/o \textsc{te} & {\color{white}{1}}3.2 16.7 38.3  &  {\color{white}{1}}3.5  17.2 38.5 &  {\color{white}{1}}5.4 22.3 45.5 &  {\color{white}{1}}5.2 22.0 45.5 &  {\color{white}{1}}2.4 13.9 35.3 &    {\color{white}{1}}2.2 13.7 35.3 \\
		\textsc{Hman} w/o \textsc{re} & 50.2 78.4 86.5 & 49.3  78.6 87.0  & 52.6 81.6 89.1   &  52.4 81.1 89.8 & 52.7  84.2 91.4 	    & 52.0 83.9 91.1 \\
		\textsc{Hman} w/o \textsc{ae}    & {49.2} {81.0} {89.8}      &  48.8 80.9 90.0    & {52.2} {83.3} {91.6}      &51.5 83.1 91.6     & {52.3} {85.6} {93.7}    & 52.3 85.1 93.2 \\
		\textsc{Hman} w/o \textsc{hw} & 46.8 76.1 84.1 &  46.0 76.2 84.6 & 50.5  79.5  87.5   &49.9 79.1 87.5 & 51.9 82.7  90.9 & 51.6  82.5  90.6 \\
		\cmidrule(l){1-7}
		
		&\multicolumn{7}{c}{\multirow{1}{*}{DBP100K}} \\
		\cmidrule(l){2-7}
		\citet{hao2016joint}         &  {\color{white}{1}}-{\color{white}{.1}} 16.9 {\color{white}{1}}-{\color{white}{.1}}       &{\color{white}{1}}-{\color{white}{.1}} 16.6 {\color{white}{1}}-{\color{white}{.1}}&  {\color{white}{1}}-{\color{white}{.1}} 21.1 {\color{white}{1}}-{\color{white}{.1}}&  {\color{white}{1}}-{\color{white}{.1}} 20.9 {\color{white}{1}}-{\color{white}{.1}} & {\color{white}{1}}-{\color{white}{.1}} 22.9 {\color{white}{1}}-{\color{white}{.1}}& {\color{white}{1}}-{\color{white}{.1}} 22.6 {\color{white}{1}}-{\color{white}{.1}}\\
		\citet{chen2017multilingual}        & {\color{white}{1}}-{\color{white}{.1}} 34.3 {\color{white}{1}}-{\color{white}{.1}}     & {\color{white}{1}}-{\color{white}{.1}} 29.1 {\color{white}{1}}-{\color{white}{.1}}&{\color{white}{1}}-{\color{white}{.1}} 33.9 {\color{white}{1}}-{\color{white}{.1}} & {\color{white}{1}}-{\color{white}{.1}} 27.2 {\color{white}{1}}-{\color{white}{.1}}      & {\color{white}{1}}-{\color{white}{.1}} 44.8 {\color{white}{1}}-{\color{white}{.1}} & {\color{white}{1}}-{\color{white}{.1}} 39.1 {\color{white}{1}}-{\color{white}{.1}} \\
		\citet{sun2017cross}   & 20.2 41.2 58.3 &  19.6 39.4 56.0   & 19.4 42.1 60.5 &  19.1 39.4 55.9  & 26.2 54.6 70.5 &  25.9  51.3  66.9  \\
		\citet{wang2018cross}   & 23.1 47.5 63.8 &  19.2 40.3 55.4  &  26.4 55.1 70.0  & 21.9 44.4 56.6 &   29.2 58.4 68.7 & 25.7 50.5  59.8  \\
		\hdashline
		\textsc{Man} & 27.2 54.2 \textbf{72.8} &  24.7 50.2 \textbf{69.0}  & 30.0 60.4 \textbf{77.3} &  26.6 54.4 {71.2} &  31.6 64.0 77.3 &   28.8 59.3 73.4    \\
    	\textsc{Man} w/o \textsc{te} & 11.8 28.6 47.7 &  11.2 28.3 47.9 & {\color{white}{1}}7.4 21.7 39.4  &  {\color{white}{1}}7.2 21.6 39.8 & {\color{white}{1}}5.4 19.4 38.2 &  {\color{white}{1}}5.1  18.8 37.1 \\
		\textsc{Man} w/o \textsc{re}  & 26.5 53.4 72.1 &  23.9 49.2 67.9   & 29.8 60.3 77.1& 26.3  53.9 70.6  & 31.0  63.2 76.4 &  28.4 58.4  72.2  \\
		\textsc{Man} w/o \textsc{ae}  & 25.5 51.7  70.4 &  22.8 47.6 66.3   & 29.4 59.4 76.1 &  25.9 52.9  69.7  & 30.8 62.7 75.8  & 28.1 57.8 71.5  \\
		\hdashline
		\textsc{Hman} & \textbf{29.8} \textbf{54.6} 69.5  &  \textbf{28.7} \textbf{53.3} \textbf{69.0}	&\textbf{34.3} \textbf{63.3}  76.1 &  \textbf{33.8} \textbf{63.0} \textbf{76.7} 	&\textbf{37.5} \textbf{67.7}  \textbf{77.7}&  \textbf{37.6} \textbf{68.1}  \textbf{78.5}\\
		\textsc{Hman} w/o \textsc{te} & {\color{white}{1}}6.8 20.3 39.2 & {\color{white}{1}}7.2 21.0 39.4 & {\color{white}{1}}3.0	11.5 27.3  & {\color{white}{1}}3.3	11.8		28.0 & {\color{white}{1}}0.5	{\color{white}{1}}3.5	11.1 &	{\color{white}{1}}0.5	{\color{white}{1}}3.4	11.4 \\
		\textsc{Hman} w/o \textsc{re} & 28.0 50.3 62.3 &  28.2 50.6  62.9  & 30.3 54.9 64.8 &  30.2 55.9 66.9&	32.8   60.3 69.1  & 33.3 60.9  69.8   \\
		\textsc{Hman} w/o \textsc{ae} & 25.7 46.4 57.3  &  25.5 64.7 57.9 &29.6 55.1 66.1 & 29.9 56.1 67.4&32.5 59.2  67.8 & 32.9  59.4 68.4 \\
		\textsc{Hman} w/o \textsc{hw} & 25.2 46.0 57.9  &  25.2 45.9 57.9 & 28.6 52.6  62.2&  28.5 53.0 63.0 & 32.8 60.9  70.0 &  32.9  60.2 70.3  \\
		\bottomrule
\end{tabular}}
\caption{Results of using graph information on DBP15K and DBP100K. @1, @10 and @50 refer to Hits@1, Hits@10 and Hits@50, respectively.}
\label{tab:results_dbp15k}
\end{table*}

\begin{table*}[t]
\centering
\scalebox{0.79}{
	\begin{tabular}{lccccccccc}
				\toprule 
		\multicolumn{1}{c}{\multirow{3}{*}{Model}}          &    \multicolumn{1}{c}{ ZH $\to$ EN  }   &  EN$\to$ ZH &   JA $\to$ EN       & EN$\to$ JA &  FR $\to$ EN & EN$\to$ FR    \\
		\cmidrule(l){2-2} \cmidrule(l){3-3}  \cmidrule(l){4-4}  \cmidrule(l){5-5} \cmidrule(l){6-6} \cmidrule(l){7-7} 	 
		& @1 {\color{white}{.}}@10 @50 & @1 {\color{white}{.}}@10 @50 & @1 {\color{white}{.}}@10 @50 & @1 {\color{white}{.}}@10 @50  & @1 {\color{white}{.}}@10 @50 & @1 {\color{white}{.}}@10 @50 \\
		\midrule	
		&\multicolumn{7}{c}{{DBP15K}} \\
		\cmidrule(l){2-7}
		Translation$\ast$ & 55.7  67.6 74.3   & 40.3	 54.2 62.2    & 74.6 84.5 89.1  &  61.9 72.0 77.2  &{\color{white}{1}}-{\color{white}{.1}} {\color{white}{1}}-{\color{white}{.1}} {\color{white}{1}}-{\color{white}{.1}}  &{\color{white}{1}}-{\color{white}{.1}} {\color{white}{1}}-{\color{white}{.1}} {\color{white}{1}}-{\color{white}{.1}}  \\
		
		JAPE + Translation$\ast$ & 73.0 90.4 96.6  &  62.7 85.2 94.2	    & 82.8 94.6 98.3   &  75.9 90.7 96.0     &{\color{white}{1}}-{\color{white}{.1}} {\color{white}{1}}-{\color{white}{.1}} {\color{white}{1}}-{\color{white}{.1}}  &{\color{white}{1}}-{\color{white}{.1}} {\color{white}{1}}-{\color{white}{.1}} {\color{white}{1}}-{\color{white}{.1}} \\
		\textsc{PairwiseBert} & 74.3 94.6 98.8	    & 74.8 94.7  99.0	    & 78.6 95.8 98.5   &  78.3 95.4	 98.4	   & 95.2  99.2 99.6 &  94.9 99.2  99.7 \\
		\hdashline
		\textsc{Man} (\textsc{Rerank}) & 84.2 93.6 94.8  &  82.1 91.8 93.1 & 89.4 94.0 94.8 &  88.2 93.3 94.0 & 93.1  95.2 95.4 & 93.1 95.3 95.4  \\
		\textsc{Hman} (\textsc{Rerank}) & 86.5 95.9 96.9  &  85.8	 94.1  95.3	    & 89.0  96.0 97.3   &  89.0	 96.0 97.5	    & 95.3 97.7 97.8   &  95.2	 97.9  98.1\\
		\textsc{Man} (\textsc{Weighted}) & 85.4 98.2 99.7 &  83.8 97.7 99.5 & 90.8 98.8 99.7 &  89.9 98.5 99.5  & 96.8 99.6 99.8 &  96.7 99.7  \textbf{99.9} \\
		\textsc{Hman} (\textsc{Weighted})    & \textbf{87.1}  \textbf{98.7}	\textbf{99.8}    & \textbf{86.4} \textbf{98.5}  \textbf{99.8}	    & \textbf{93.5} \textbf{99.4}	\textbf{99.9}     &  \textbf{93.3} \textbf{99.3}	 \textbf{99.9}	    & \textbf{97.3} \textbf{99.8} \textbf{99.9} 	    &  \textbf{97.3} \textbf{99.8}  \textbf{99.9} \\
		\midrule
		&\multicolumn{9}{c}{\multirow{1}{*}{DBP100K}} \\
		\cmidrule(l){2-10}
		\textsc{PairwiseBert} & 65.1 85.1  92.6	& 66.2  85.8  92.9	& 67.7  86.5  93.1 	&  67.9 86.4 93.2	& 93.2 97.9 98.9	& 93.4 98.0  98.9\\
		\hdashline			
		\textsc{Man} (\textsc{Rerank}) & 59.5 62.1 62.2 &   55.9 58.2 58.2 & 65.5  68.2 68.4  &59.9 62.1  62.3& 69.7  70.4  70.5 &  65.5 66.2 66.2  \\
		\textsc{Hman} (\textsc{Rerank}) & 58.9 61.2  61.3 &  57.9 60.2 60.3 & 66.9 69.4 69.6 &  67.0 69.6  69.8& 72.1 72.9   73.0&   72.7 73.5 73.5 \\
		\textsc{Man} (\textsc{Weighted}) &\textbf{81.4} \textbf{94.9} \textbf{98.2} &  \textbf{80.5} 94.1 97.7 & 84.3  {95.4}  98.3 &  81.5 {94.2} 97.6 & 96.2 {99.3} \textbf{99.7} & 95.7  99.1  99.6 \\ 
		\textsc{Hman} (\textsc{Weighted})  & 81.1 94.3  97.8	&  80.3 \textbf{94.5} \textbf{97.9}	& \textbf{85.2} \textbf{96.1}  \textbf{98.4}	&  \textbf{84.6} \textbf{96.1} \textbf{98.5}	& \textbf{96.5} \textbf{99.4} \textbf{99.7}	&  \textbf{96.5} \textbf{99.5}	\textbf{99.8}\\
		\bottomrule
\end{tabular}}
\caption{Results of using both graph and textual information on DBP15K and DBP100K. @1, @10, and @50 refer to Hits@1, Hits@10, and Hits@50, respectively. $\ast$ indicates results are taken from \cite{sun2017cross}. }
\label{tab:text_result_dbp15k}
\end{table*}

\subsection{Results on Graph Embeddings}
\label{sec:result-gcn}
We first compare \textsc{Man} and \textsc{Hman} against previous systems \cite{hao2016joint,chen2017multilingual,sun2017cross,wang2018cross}.
As shown in Table \ref{tab:results_dbp15k}, \textsc{Man} and \textsc{Hman} consistently outperform all baselines in all scenarios, especially \textsc{Hman}. 
It is worth noting that, in this case, \textsc{Man} and \textsc{Hman} use as much information as \citet{wang2018cross}, while \citet{sun2017cross} require extra supervised information (relations and attributes of two KGs need to be aligned in advance).
The performance improvements confirm that our model can better utilize topological, relational, and attribute information of entities provided by KGs.

To explain why \textsc{Hman} achieves better results than \textsc{Man}, recall that \textsc{Man} collects relation and attribute information by the propagation mechanism in GCNs where such knowledge is exchanged through neighbors, while \textsc{Hman} uses feedforward networks to capture expressive features directly from the input feature vectors without propagation.
As we discussed before, it is not always the case that neighbors of equivalent entities share similar relations or attributes.
Propagating such features through linked entities in GCNs may introduce noise and thus harm performance.

Moreover, we perform ablation studies on the two proposed models to investigate the effectiveness of each component.
We alternatively remove each aspect of features (i.e., topological, relation, and attribute features) and the highway layer in \textsc{Hman}, denoted as w/o \textsc{te} (\textsc{re}, \textsc{ae}, and \textsc{hw}).
As reported in Table \ref{tab:results_dbp15k}, 
we observe that after removing relation or attribute features, the performance of \textsc{Hman} and \textsc{Man} drops across all datasets.
These figures prove that these two aspects of features are useful in making alignment decisions. 
On the other hand, compared to \textsc{Man}, \textsc{Hman} shows more significant performance drops, which also demonstrates that employing the feedforward networks can better categorize relation and attribute features than GCNs in this scenario.
Interestingly, looking at the two variants \textsc{Man} w/o \textsc{te} and \textsc{Hman} w/o \textsc{te},
we can see the former achieves better results.
Since \textsc{Man} propagates relation and attribute features via graph structures, it can still implicitly capture topological knowledge of entities even after we remove the topological features.
However, \textsc{Hman} loses such structure knowledge when topological features are excluded, and thus its results are worse. 
From these experiments, we can conclude that the topological information is playing an indispensable role in making alignment decisions.

\subsection{Results with Textual Embeddings}

In this section, we discuss empirical results involving the addition of entity descriptions, shown in Table \ref{tab:text_result_dbp15k}.
Applying literal descriptions of entities to conduct cross-lingual entity alignment is relatively under-explored.
The recent work of \citet{chen2018co} used entity descriptions in their model; 
however, we are unable to make comparisons with their work, as we do not have access to their code and data.
Since we employ BERT to learn textual embeddings of descriptions, we consider systems that also use external resources, like Google Translate,\footnote{\url{https://cloud.google.com/translate/}} as our baselines.
We directly take results reported by \citet{sun2017cross}, denoted as ``Translation'' and ``JAPE$+$Translation''.

The \textsc{PointwiseBert} model is used with GCN-based models, which largely reduces the search space, as indicated by \textsc{Man} (\textsc{Rerank}) and \textsc{Hman} (\textsc{Rerank}), where the difference is that the candidate pools are given by \textsc{Man} and \textsc{Hman}, respectively.
For DBP15K, we select top-200 candidate target entities as the candidate pool while for DBP100K, top-20 candidates are selected due to its larger size.
The reranking method does lead to performance gains across all datasets, where the improvements are dependent on the quality of the candidate pools.
\textsc{Hman} (\textsc{Rerank}) generally performs better than \textsc{Man} (\textsc{Rerank}) since \textsc{Hman} recommends more promising candidate pools. 

The \textsc{PairwiseBert} model learns the textual embeddings that map cross-lingual descriptions into the same space,
which can be directly used to align entities. The results are listed under \textsc{PairwiseBert} in Table \ref{tab:text_result_dbp15k}.
We can see that it achieves good results on its own, which also shows the efficacy of using multilingual descriptions. 
Moreover, such textual embeddings can be combined with graph embeddings (learned by \textsc{Man} or \textsc{Hman}) by weighted concatenation, as discussed in Section \ref{sec:strategy}.
The results are reported as \textsc{Man} (\textsc{Weighted}) and \textsc{Hman} (\textsc{Weighted}), respectively.
As we can see, this simple operation leads to significant improvements and gives excellent results across all datasets.
However, it is not always the case that KGs provide descriptions for every entity.
For those entities whose descriptions are not available, the graph embeddings would be the only source for making alignment decisions.

\begin{table}[t]
\centering
\scalebox{0.6}{
\begin{tabular}{p{4em} p{13em} p{12em} }        \toprule
&\textbf{English} & \textbf{Chinese} \\  \midrule
\textbf{ILL pair} & Casino\_Royale\_(2006\_film) (3)  &  007\begin{CJK}{UTF8}{min}大戰皇家賭場\end{CJK} (3)   \\  
\textbf{Features}   & \blue{starring, starring}, distributor & \blue{starring, starring}, language \\
\textbf{Neighbors} & \vtop{\hbox{\strut Daniel\_Craig (1), Eva\_Green (4),}\hbox{\strut \red{Columbia\_Pictures (9)}}} & \begin{CJK}{UTF8}{min}丹尼爾·克雷格\end{CJK} (1), \begin{CJK}{UTF8}{min}伊娃·格蓮\end{CJK} (4), \red{\begin{CJK}{UTF8}{min}英語\end{CJK} (832)}\\
\bottomrule
\end{tabular}}
\caption{Case study of the noise introduced by the propagation mechanism.}
\label{tab:error-analysis}
\end{table}

\subsection{Case Study}

In this section, we describe a case study to understand the performance gap between \textsc{Hman} and \textsc{Man}. 
The example in Table~\ref{tab:error-analysis} provides insights potentially explaining this performance gap.
We argue that \textsc{Man} introduces unexpected noise from heterogeneous nodes during the GCN propagation process.
We use the number in parentheses (*) after entity names to denote the number of relation features they have. 

In this particular example, the two entities ``\emph{Casino\_Royale\_(2006\_film)}'' in the source language (English) and ``\emph{007\begin{CJK}{UTF8}{min}大戰皇家賭場\end{CJK}}'' in the target language (Chinese) both have three relation features.
We notice that the propagation mechanism introduces some neighbors which are unable to find cross-lingual counterparts from the other end, marked in red.
Considering the entity ``\emph{\begin{CJK}{UTF8}{min}英語\end{CJK}}''~(English), a neighbor of ``\emph{007\begin{CJK}{UTF8}{min}大戰皇家賭場\end{CJK}}'', no counterparts can be found in the neighbors of ``\emph{Casino\_Royale\_(2006\_film)}''.
We also observe that ``\emph{\begin{CJK}{UTF8}{min}英語\end{CJK}}''~(English) is a pivot node in the Chinese KG and has 832 relations, such as 
``\begin{CJK}{UTF8}{min}語言\end{CJK}''~(Language), ``\begin{CJK}{UTF8}{min}官方語言\end{CJK}''~(Official Language), and ``\begin{CJK}{UTF8}{min}頻道語言\end{CJK}''~(Channel Language).
In this case, propagating features from neighbors can harm performance. 
In fact, the feature sets of the ILL pair already convey information that captures their similarity (e.g., the ``starring'' marked in blue are shared twice).
Therefore, by directly using feedforward networks, \textsc{Hman} is able to effectively capture such knowledge.

\section{Related Work}
\label{sec:related_work}

\textbf{KG Alignment.}
Research on KG alignment can be categorized into two groups:\ monolingual and multilingual entity alignment.
As for monolingual entity alignment, main approaches align two entities by computing string similarity of entity labels \cite{scharffe2009rdf,volz2009discovering,ngomo2011limes} or graph similarity \cite{raimond2008automatic,pershina2015holistic,azmy2019matching}.
Recently, \citet{trsedya2019entity} proposed an embedding-based model that incorporates attribute values to learn the entity embeddings.

To match entities in different languages, \citet{wang12cross} leveraged only language-independent information to find possible links cross multilingual Wiki knowledge graphs.
Recent studies learned cross-lingual embeddings of entities based on TransE \cite{bordes2013translating}, which are then used to align entities across languages.
\citet{chen2018co} designed a co-training algorithm to alternately learn multilingual entity and description embeddings.
\citet{wang2018cross} applied GCNs with the connectivity matrix defined on relations to embed entities from multilingual KGs into a unified low-dimensional space.

In this work, we also employ GCNs.
However, in contrast to \citet{wang2018cross}, we regard relation features as input to our models.
In addition, we investigate two different ways to capture relation and attribute features.

\smallskip
\noindent
\textbf{Multilingual Sentence Representations.}
Another line of research related to this work is aligning sentences in multiple languages.
Recent works \cite{hermann2014multilingual,conneau2018xnli,eriguchi2018zero} studied cross-lingual sentence classification via zero-shot learning.
\citet{johnson2017google} proposed a sequence-to-sequence multilingual machine translation system where the encoder can be used to produce cross-lingual sentence embeddings \cite{artetxe2018massively}.
Recently, BERT \cite{devlin2019bert} has advanced the state-of-the-art on multiple natural language understanding tasks.
Specifically, multilingual BERT enables learning representations of sentences under multilingual settings.
We adopt BERT to produce cross-lingual representations of entity literal descriptions to capture their semantic relatedness, which benefits cross-lingual entity alignment.

\section{Conclusion and Future Work}
In this work, we focus on the task of cross-lingual entity alignment, which aims to discover mappings of equivalent entities in multilingual knowledge graphs.
We proposed two GCN-based models and two uses of multilingual BERT to investigate how to better utilize multi-aspect information of entities provided by KGs, including topological connections, relations, attributes, and entity descriptions.
Empirical results demonstrate that our best model consistently achieves state-of-the-art performance across all datasets.
In the future, we would like to apply our methods to other multilingual datasets such as YAGO and BabelNet. 
Also, since literal descriptions of entities are not always available, we will investigate alternative ways to design graph-based models that can better capture structured knowledge for this task.

\section*{Acknowledgments}
We would like to thank the three anonymous reviewers for their thoughtful and constructive comments.
Waterloo researchers are supported by the Natural Sciences and Engineering Research Council (NSERC) of Canada, with additional computational resources provided by Compute Ontario and Compute Canada.
Yanyan Zou and Wei Lu are supported by Singapore Ministry of Education Academic Research Fund (AcRF) Tier 2 Project MOE2017-T2-1-156, and are partially supported by SUTD project PIE-SGP-AI-2018-01.

\bibliography{emnlp2019}
\bibliographystyle{acl_natbib}

\end{document}